\documentclass[sigconf]{acmart}

\usepackage{balance}

\usepackage{booktabs} 
\usepackage[utf8]{inputenc}
\usepackage{epsfig}
\usepackage{graphicx}
\usepackage{subcaption}
\usepackage{caption}
\usepackage{amsmath}
\usepackage{hyperref}
\usepackage{tablefootnote}
\usepackage{multirow}
\usepackage[autostyle]{csquotes}
\usepackage{numprint}
\usepackage{adjustbox}

\title{MAVE: A Product Dataset for Multi-source Attribute Value Extraction}

\author{Li Yang$^{*,1}$, Qifan Wang$^{*,1}$, Zac Yu$^2$, Anand Kulkarni$^2$, Sumit Sanghai$^1$, Bin Shu$^2$, Jon Elsas$^2$, Bhargav Kanagal$^1$}
\thanks{$^{*}$ Equal contributions.}
\affiliation{%
  \institution{$^1$Google Research, Mountain View, USA \ \ $^2$Google, Pittsburgh, USA}
  \country{}
}
\email{{lyliyang, wqfcr, zacyu, anandkulkarni, sumitsanghai, bins, jelsas, bhargav}@google.com}

\copyrightyear{2022}
\acmYear{2022}
\setcopyright{acmcopyright}\acmConference[WSDM '22]{Proceedings of the Fifteenth
ACM International Conference on Web Search and Data Mining}{February 21--25,
2022}{Tempe, AZ, USA}
\acmBooktitle{Proceedings of the Fifteenth ACM International Conference on Web
Search and Data Mining (WSDM '22), February 21--25, 2022, Tempe, AZ, USA}
\acmPrice{15.00}
\acmDOI{10.1145/3488560.3498377}
\acmISBN{978-1-4503-9132-0/22/02}

\begin{document}
\sloppy\hyphenpenalty=3500
\fancyhead{}

\begin{abstract}
Attribute value extraction refers to the task of identifying values of an attribute of interest from product information. Product attribute values are essential in many e-commerce scenarios, such as customer service robots, product ranking, retrieval and recommendations. While in the real world, the attribute values of a product are usually incomplete and vary over time, which greatly hinders the practical applications. In this paper, we introduce MAVE, a new dataset to better facilitate research on product attribute value extraction. MAVE is composed of a curated set of 2.2 million products from Amazon pages, with 3 million attribute-value annotations across 1257 unique categories. MAVE has four main and unique advantages: First, MAVE is the largest product attribute value extraction dataset by the number of attribute-value examples. Second, MAVE includes multi-source representations from the product, which captures the full product information with high attribute coverage. Third, MAVE represents a more diverse set of attributes and values relative to what previous datasets cover. Lastly, MAVE provides a very challenging zero-shot test set, as we empirically illustrate in the experiments. We further propose a novel approach that effectively extracts the attribute value from the multi-source product information. We conduct extensive experiments with several baselines and show that MAVE is an effective dataset for attribute value extraction task. It is also a very challenging task on zero-shot attribute extraction. Data is available at {\it \url{https://github.com/google-research-datasets/MAVE}}.
\end{abstract}


\maketitle

\begin{figure}[th]
\begin{center}
\includegraphics[width=0.85\linewidth]{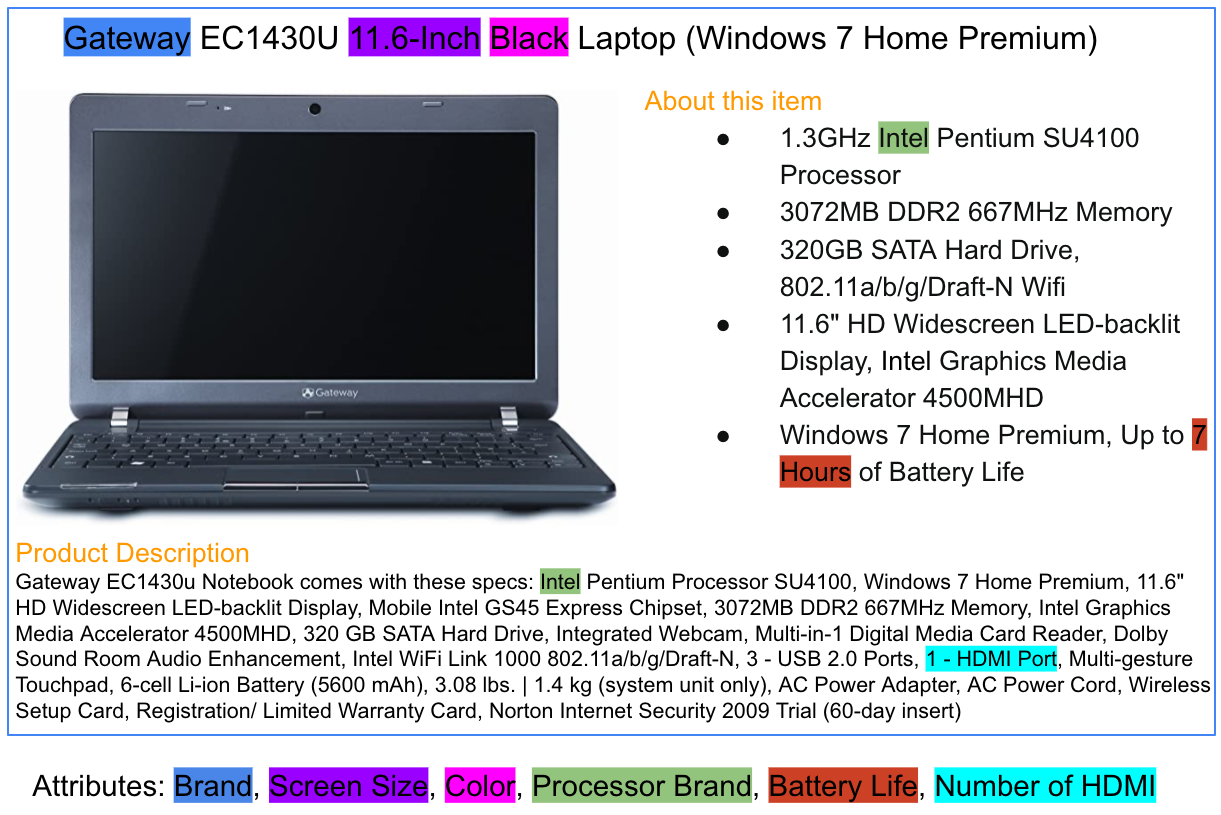}
\end{center}
\caption{An example of the product profile on an e-Commerce platform. It consists of multiple information sources, including a product title, a product description and several other sources.} \label{fig:example}
\end{figure}
\section{Introduction}
Product attributes are critical product features which form an essential component of e-commerce platforms today.
They provide useful details of the product, which help customers to make purchasing decisions and facilitate retailers on many applications, such as product search ~\cite{Search1,Search2}, product recommendation ~\cite{Rec1,Rec2}, product representation ~\cite{Embedding1,Embedding2} and question answering system ~\cite{QA1,QA2}.
However, for most retailers, product attributes are often noisy and incomplete with a lot of missing values, which has been shown in many previous studies ~\cite{ZhuWLWHZ20, WangYKSSSYE20}.
Therefore, it is an important research problem to supplement the product with missing values for attributes of interest.
The problem of attribute value extraction is illustrated in Figure \ref{fig:example}. It shows
the profile of a Laptop product as an example, which consists of a title, a product description and several other information sources. It also
shows several attribute values that could be extracted.

There has been a lot of interest in this topic, and a plethora of research ~\cite{GhaniPLKF06,ProbstGKFL07,thesis17,CarmelLM18,RezkANZ19,ZhaoGS19,0005FC0S19} in this area both in academia and industry. Early works on this problem are rule based approaches~\cite{ChiticariuKLRV10,VandicDF12,GopalakrishnanIMRS12}, which utilize domain-specific knowledge to design regular expressions. 
Several other approaches such as ~\cite{PutthividhyaH11,More16,YanGDGZQ20} formulate the attribute value extraction as an instance of named entity recognition (NER) problem \cite{nadeau2007survey}, and build extraction models to identify the entities/values from the input text.
With the recent advance in natural language understanding, sequence tagging~\cite{ZhengMD018,XuWMJL19,WangYKSSSYE20,Amazon2} based approaches have been proposed, which achieve promising results.
Attribute value extraction approaches rely on a rich dataset to help them learn to extract the attribute values from the product profile. Existing attribute datasets are created from different e-commerce platforms, such as Amazon ~\cite{Amazon1,Amazon2}, AliExpress ~\cite{XuWMJL19,WangYKSSSYE20}, JD ~\cite{ZhuWLWHZ20}, etc.
However, there are several major limitations:
\begin{itemize}
\item The scales of current public datasets are not large. In fact, many of the datasets used in previous works are not even publicly available. Moreover, existing datasets, such as the MAE one in Table \ref{table:datasets}, are very noisy and cannot be used without dense pre-processing and cleanup. Researches have shown that having a large and accurate attribute-value dataset can significantly improve model performance.
\item The product profiles are relatively simple. Most product context is formed from a single source, i.e., product title or description, resulting in short sequences. In real world applications, products are usually represented by multiple sources, such as title, description, specification, table, review, image, etc, forming much longer sequences.
\item The attributes and values are not diverse. The lack of the data diversity impedes research in the zero-shot/few-shot attribute value extraction, which has huge potential impact on unseen data. With the rapid expansion of e-commerce, new products with new capabilities are being released constantly which means the model needs to gracefully adapt to new attributes. 
\end{itemize}

To address these challenges and advance research on product attributes, we present the Multi-source Attribute Value Extraction (MAVE) dataset.
MAVE is created by first extracting the raw values from Amazon Review pages~\cite{AmazonReview} on a pre-defined set of attributes. 
We then apply rigorous filtering and post-mapping to only retain high quality extractions. The resulting dataset contains over
2.2 million products distributed into 1257 different product categories, with 3 million attribute-value annotations - making MAVE the largest attribute value extraction dataset at the time of writing.
Moreover, MAVE provides more complete product profiles from multiple information sources, including product title, descriptions, features, specifications and key-value pairs, which enhance the attribute coverage. For example, in Figure \ref{fig:example}, the value {\it 7 Hours} for the attribute {\it Battery Life} is from the product specifications.
Furthermore, MAVE contains a rich and diverse set of attributes and values. In particular, there are 2535 unique attributes with 100K unique values in the MAVE dataset.
It is worth pointing out that by leveraging the human rules based matching, MAVE is able to ensure a high quality bar with 97.8\% precision (manual check) over 1000 random sampled attribute-value annotations.

In this paper, we also propose a novel Multi-source Attribute Value Extraction model via Question Answering (MAVEQA). MAVEQA uses the recent ETC ~\cite{ETC} model as the backbone to effectively extract the attribute value from the multiple information sources with the long sequences.
We conduct an extensive set of experiments on the MAVE dataset over several state-of-the-art baselines (including the proposed MAVEQA). The experimental results demonstrate the potency of the data. They also show that MAVE is a challenging dataset on attribute value extraction for various attributes, especially those zero-shot attributes.

\section{Related Work}
\subsection{Attribute Value Extraction Models}
Rule-based extraction methods ~\cite{nadeau2007survey,VandicDF12,GopalakrishnanIMRS12} are among those early works on attribute value extraction. They use a domain-specific seed dictionary or vocabulary to identify key phrases and attributes. Ghani et al. \cite{GhaniPLKF06} predefine a set of attributes to extract the corresponding values. Wong et al. \cite{WongWLN09} extract attribute-value pairs from the semi-structured text. Shinzato and Sekine \cite{ShinzatoS13} design an unsupervised method to extract attribute values from product description. Several rule-based and linguistic approaches ~\cite{MikheevMG99,ChiticariuKLRV10} leverage syntactic structure of sentences to extract dependency relations. 
Several NER based systems ~\cite{PutthividhyaH11,CollobertWBKKK11,LingW12,More16} were proposed for extracting product attributes and values. However, these rule-base and domain-specific methods suffer from limited coverage and closed world assumptions.
\begin{table}
\small
\begin{center}
\resizebox{0.4\textwidth}{!}{\begin{tabular}{|c|c|c|c|c|c|}
\hline
Dataset & \#products  & \#annotations  & \#sources  & public\\
\hline
MAE ~\cite{LoganHS17}& 600K &  2.2M & 2 & Yes\\
\hline
OpenTag ~\cite{ZhengMD018}& 10K &  13K & 2 & No\\
\hline
MEPAVE ~\cite{ZhuWLWHZ20} & 34K &  87K & 2 & Yes\\
\hline
AE-110K ~\cite{XuWMJL19} & 50K  &  110K & 1 & Yes\\
\hline
AdaTag ~\cite{Amazon2} & 333K  &  410K & Multiple & No\\
\hline
MAVE & 2.2M &  3M & Multiple & Yes \\
\hline
\end{tabular}}
\end{center}
\caption{Existing datasets for attribute value extraction.}\label{table:datasets}
\end{table}

With the development of deep neural networks, various neural network methods have been proposed and applied in sequence tagging successfully.
Huang et al. \cite{HuangXY15} are the first to apply the BiLSTM-CRF model to a sequence tagging task. But it employs heavy feature engineering to extract character-level features.
Lample et al. \cite{LampleBSKD16} utilize BiLSTM to model both word-level and character-level information rather than hand-crafted features.
Chiu and Nichols \cite{ChiuN16} model character level information using convolutional neural network (CNN), which achieves competitive performance and has been extended in \cite{MaH16} with an end to end LSTM-CNNs-CRF model.

Recently, several approaches employ sequence tagging models for attribute value extraction.
Kozareva et al. \cite{KozarevaLZG16} adopt BiLSTM-CRF model to tag several product attributes from search
queries with hand-crafted features. Furthermore, Zheng et al. \cite{ZhengMD018} develop an end-to-end tagging model utilizing BiLSTM, CRF, and attention mechanism without any dictionary. More recently, Xu et al. \cite{XuWMJL19} adopt only one global set of BIO tags for any attributes to scale up the model, which explicitly models the semantic representations for attribute and product title. 
Most recently, several approaches ~\cite{WangYKSSSYE20,LiuZSFX20,Amazon2} formulate the problem as a question answering (QA)  ~\cite{QAReview,QAReview2} task. Wang et al. \cite{WangYKSSSYE20} propose to jointly encode both the attribute and the product context with cross-attention model, and then decode the value span with a BIO tagging.
It is worth mentioning that there are also methods that work on multimodal attribute value extraction ~\cite{LoganHS17,ZhuWLWHZ20,Amazon1}, which incorporate the visual features using CNN \cite{LiuCYMYC21} to boost coverage of the attributes.

Relation extraction ~\cite{PetrovskiBB14,PengPQTY17,RiedelYM10,SuchanekIW06,ZengLLZZ14,WangKGS19} is also relevant to attribute value extraction. Relation extraction refers to the task of extracting relational tuples and putting them in a knowledge base. The tuple usually corresponds to a subject, a predicate and an object. Attribute value extraction can be thought of as the problem where the subject is known (the product), and given the attribute (the relation) extract the value. However, relation extraction has traditionally focused on extracting relations from sentences relying on entity linking systems to identify the subject/object and building models to learn the predicates in a sentence ~\cite{LevySCZ17,0005FC0S19}. Whereas in attribute value extraction ~\cite{QiuBDSS15,PetrovskiB17}, usually the predicates (the attribute names) rarely occur in the product title or the description, and entity linking is very hard because the domain of all entities/values is unknown.

\subsection{Attribute Value Extraction Datasets}
Although product attribute value extraction is an important research problem, which has raised a lot of attention in recent years, as far as we know, there is no universal dataset for evaluating the AVE models. This is one of the motivations of this work, to establish such a benchmark dataset that could serve as a rich resource to drive research in attribute value extraction. However, there are a few datasets that are used in previous AVE approaches. MAE \cite{LoganHS17} is among the early datasets that are applied to this space. MAE is composed of mixed media data over 2 million product items, obtained by running the Diffbot Product API on over 20 million web pages from 1068 different commercial websites. The attribute-value pairs are automatically extracted from tables present on product webpages without any verification, which are very noisy - only less than 30\% values can be found in about 600K products. OpenTag \cite{ZhengMD018} is one of the pioneer work that models this problem as sequence tagging with LSTM. However, there are only five attributes in this data with 13K attribute value annotations. MEPAVE ~\cite{ZhuWLWHZ20,abs-2104-06960} is a multimodal dataset with textual product descriptions and images. It contains 34K products collected from a JD e-commerce platform. The 87K attribute value annotations are generated by crowd sourcing annotators. AE-110K \cite{XuWMJL19} is a public dataset collected from AliExpress Sports \& Entertainment category. It is composed of 110K attribute value annotations obtained from Item Specific in 50K product titles. Despite the long-tailed attribute distribution (7650 of 8906 attributes occur less than 10 times), this dataset has been used in several recent works ~\cite{WangYKSSSYE20,roy-etal-2021-attribute}. AdaTag \cite{Amazon2} builds a dataset by collecting product profiles with multiple source representations from the public web pages at Amazon.com. The 410K attribute value annotations are generated from 333K products in a similar way to the previous work ~\cite{ZhengMD018,XuWMJL19,KaramanolakisMD20}. Unfortunately, this dataset is not publicly available. There are many other works ~\cite{ECNLP2,KDD2} that use their own datasets, which are all small data with limited attributes.
We summarize the existing datasets in Table \ref{table:datasets}.
\section{MAVE: Multi-source Attribute Value Extraction Dataset}
Our MAVE dataset is derived from a public product collection - the Amazon Review Dataset ~\cite{AmazonReview}\footnote{\url{https://nijianmo.github.io/amazon/index.html}} by generating and adding attribute-value annotations to the products.
In this section, we describe our pipeline for creating MAVE, and also provide detailed statistics from various aspects. 

\subsection{Product Profiles}\label{subsection:product-text-features}
The products from the Amazon Review Dataset contain multiple information sources. For each product, we organize the `title', `description', `feature', `price', and `brand' into a structured product profile. We performed the following cleaning of the texts to ensure a high quality of the attribute value annotations.
\begin{itemize}
    \item Remove the html tags, e.g., script and style tags, within the product profile by using BeautifulSoup \footnote{\url{https://www.crummy.com/software/BeautifulSoup/bs4/doc/}} as well as a few hard-coded rules.
    \item Remove extra whitespaces and invalid text unicode characters.
    \item Remove products with a total number of words less than 20.
    \item Remove products without a title.
\end{itemize}
Table~\ref{table:product_example} shows an example of one product profile after the pre-processing.

\begin{table}
\footnotesize	
\begin{tabular}{|p{0.15\linewidth} | p{0.8\linewidth}|}
\hline
{\bf Source} & {\bf Text} \\
\hline
title & Ty \textcolor{red}{Beanie Baby} Ants the Anteater \\
\hline
description 1& Ty \textcolor{green}{Beanie Babies} Ants the Anteater. Approximately 8" long, 4" tall. Birthday: November 7, 1997 Poem: Most anteaters love to eat bugs But this little fellow gives big hugs. He'd rather dine on apple pie Than eat an ant or harm a fly. \\
\hline
description 2& This most unusual \textcolor{blue}{Beanie Baby} is brimming with personality. Ants was born November 7, 1997. His poem reads: Most anteaters love to eat bugs But this little fellow gives big hugs He'd rather dine on apple pie Than eat an ant or harm a fly! This long-snouted guy, balancing on his long tail, is just adorable. His head and tail are gray; his middle is three stripes white, black, and white; and he has sweet black button eyes. His tiny gray felt ears really give this guy some charm. Surface wash only. \\
\hline
feature 1 & Ty \textcolor{orange}{beanie baby} - Ants the anteater \\
\hline
feature 2 & Birthday: November 7, 1997 \\
\hline
feature 3 & Approx 8" long \\
\hline
price  & \$6.06 \\
\hline
brand  & TY \\
\hline
{\bf Attribute} & {\bf Values} \\
\hline
Toy Maker & \{\textcolor{red}{Beanie Baby}, title, span (3, 13)\}, \ \{\textcolor{green}{Beanie Babies}, description 1, span (3, 15)\}, \  \{\textcolor{blue}{Beanie Baby}, description 2, span (18, 28)\}, \  \{\textcolor{orange}{beanie baby}, feature 1, span (3, 13)\}\\
\hline
\end{tabular}
\caption{An example of a product profile with annotations. Each source corresponds to a field of the product.} \label{table:product_example}
\end{table}

\subsection{Product Attributes}
Before describing how to generate and associate attribute values to product profiles, we first discuss the structure of the attributes. 

\subsubsection{Attributes are category specific} Products can be classified into different categories, e.g. {\it Shoes}, {\it Books}, {\it Mobile Phones}. For each category, a set of attributes are pre-defined by our human raters. For example, the {\it Shoes} category contains attributes like {\it Type}, {\it Style}, {\it Heel Height}, etc.; The {\it Books} category has attributes {\it Type}, {\it Format}, {\it Fiction Form}, etc.; The {\it Mobile Phones} category has attributes {\it Operating System}, {\it Battery Life}, {\it Screen Size}, etc. It is worth pointing out that certain attributes are more general, which are shared across multiple categories, e.g., {\it Type}, {\it Style}. But some attributes are more specific and apply to only a few categories, e.g. {\it Battery Life}, {\it Fiction Form}. Moreover, different attributes from different categories can share similar meanings, for example, {\it Resolution} for {\it Digital Cameras} and {\it Rear Camera Resolution} for {\it Mobile Phones}. The correlation among attributes empowers attribute value extraction models to transfer knowledge from one attribute to another.

\subsubsection{Attribute values are category specific}
For attributes shared across multiple categories, the definition of attribute values may also vary. For example, the {\it Type} for {\it Books} could be {\it Fiction} or {\it Non-fiction}, but the {\it Type} for {\it Shoes} could be {\it Sandals}, {\it Slippers}, etc.
Note that some of the attributes have discrete or categorical values, e.g. {\it Style}, {\it Type}, which are associated with a set of well defined attribute values. On the other side, some attributes have measure, i.e., numerical values with units, e.g. {\it Heel Height}, {\it Screen Size}.

\subsection{Attribute Value Annotations}\label{subsection:generating-attribute-labels}
In this section, we present the details of generating attribute value annotations on the product profiles. In the following discussion, we refer to an example as a tuple of (product, category, attribute).

\subsubsection{Category classification}
We first run a category classifier on a product to predict the category. The category vocabulary is defined by human experts, and the category classifier is a multi-class classification model trained on an existing large dataset containing category labels. We remove the products with predicted category probability lower than a threshold of $0.5$ to ensure a high quality of category classifications.

\begin{table}
\small
\begin{tabular}{|c|c|c|}
\hline
Counts & Positives & Negatives \\
\hline
\# products & 2226509 & 1248009 \\
\# product-attribute pairs & 2987151 & 1780428 \\
\hline
{\rm \# products with 1-2 attributes} & 2102927 & 1140561 \\
{\rm \# products with 3-5 attributes} & 121897 & 99896 \\
{\rm \# products with $\ge$6 attributes} & 1685 & 7552 \\
\hline
\# unique categories & 1257 & 1114 \\
\# unique attributes  & 705 & 693 \\
\# unique category-attribute pairs  & 2535 & 2305 \\
\hline
\end{tabular}
\caption{Overall statistics of the MAVE dataset.} \label{table:general-statistics}
\end{table}


\begin{figure}
\includegraphics[width=0.8\linewidth]{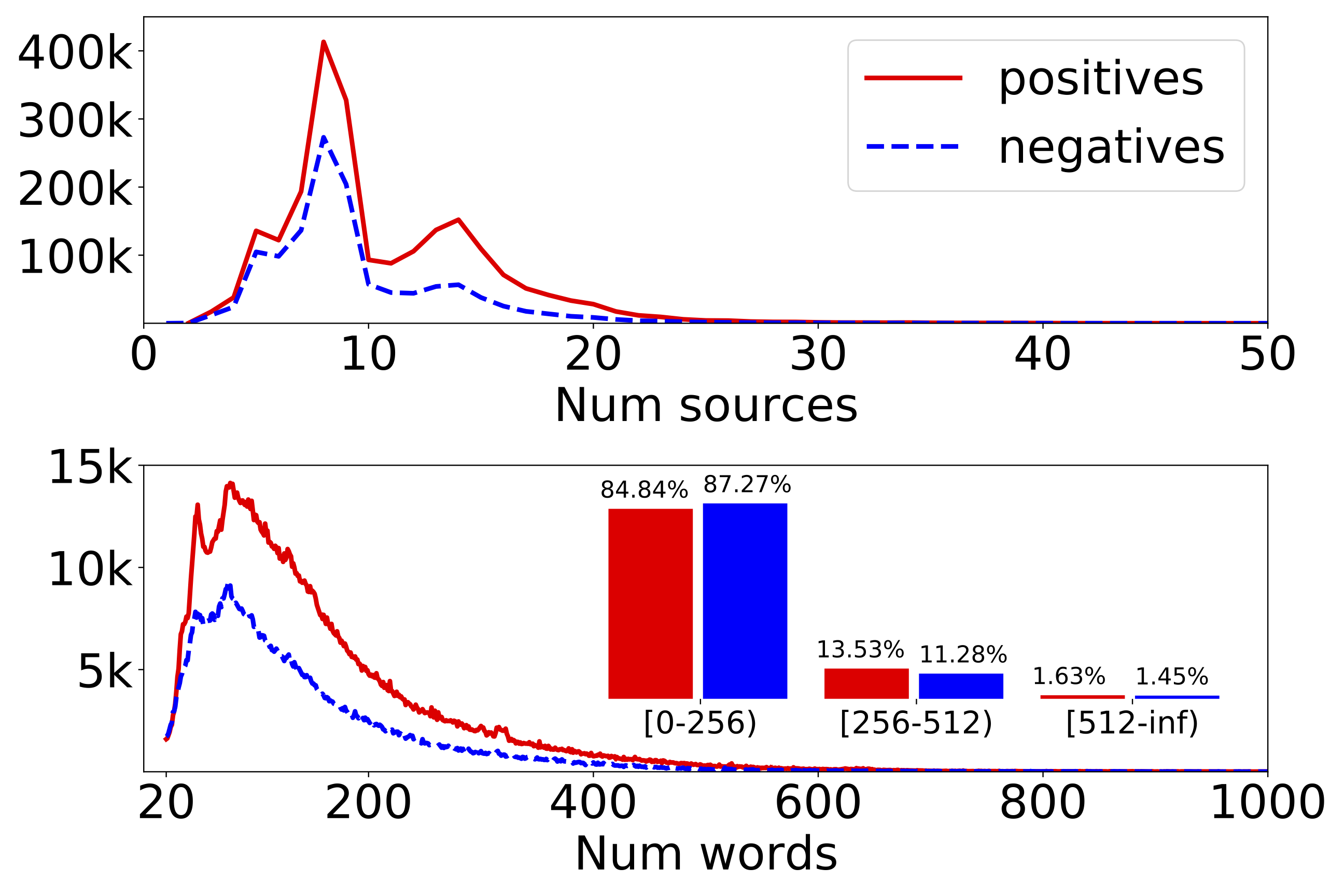}
\caption{Distributions on \# sources and \# words for positive and negative sets. The inset figure is the percentage in each \# words buckets within each sets.} \label{fig:sequence_length}
\end{figure}

\begin{figure}
\centering
\includegraphics[width=0.8\linewidth]{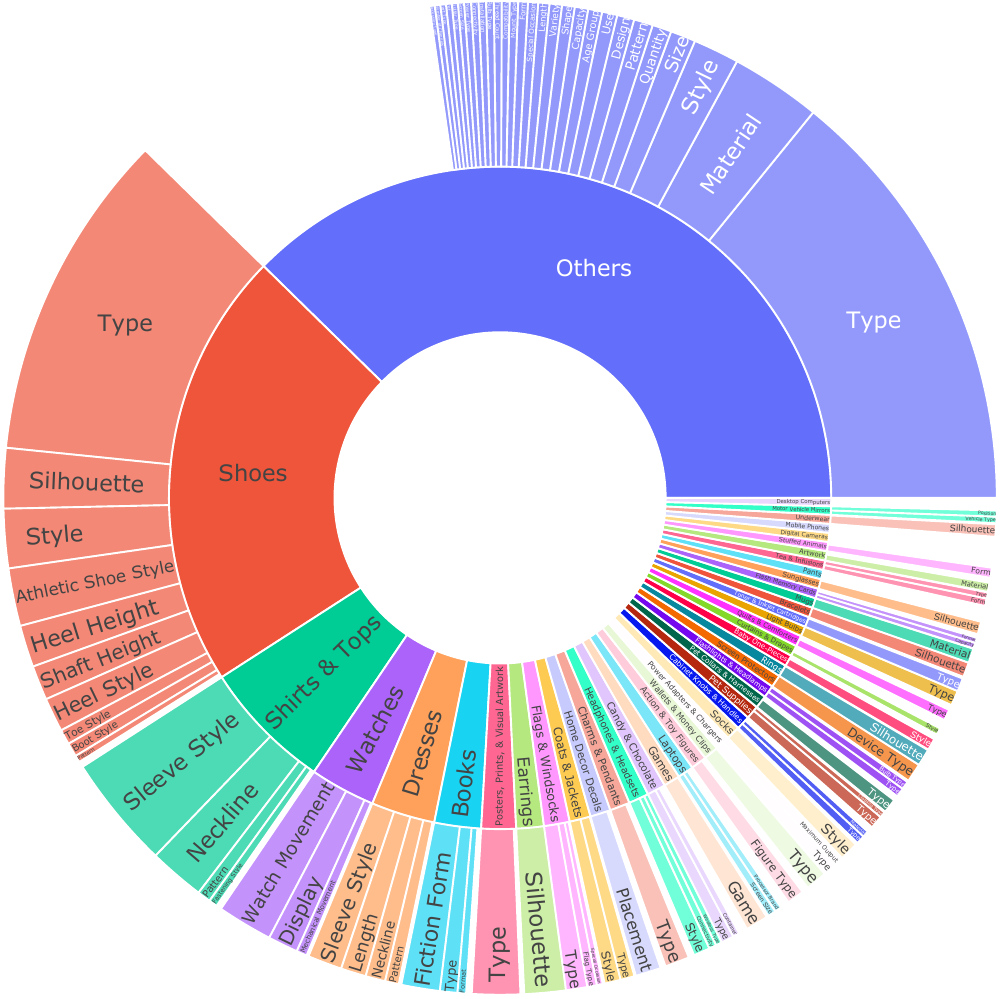}
\caption{
Number of positive example distributions on categories and attributes. Inner and outer circles are for categories and attributes, respectively. Empty blocks indicate the category-attribute contains too few examples to show.} \label{fig:attribute-statistics}
\end{figure}

\subsubsection{Attribute value span extraction}
As aforementioned, for each category, a set of attributes are pre-defined. For each attribute in a category, a set of extraction rules are also defined by our human raters to extract value spans in product profiles and map to normalized attribute values. We do not apply the extraction rules directly to the products because the extraction rules are not complete - they are designed for extracting specific formats of attribute values, resulting in low coverage extractions. Moreover, the extraction rules are not designed to extract all the attribute value spans in the context. Instead, we first train an ensemble of five versions of AVEQA~\cite{WangYKSSSYE20} models using a large amount of silver data from the extraction rules as well as human verified gold data. These models vary in terms of random initialization, training data version, pre-post processing, etc. The input of these models is a concatenation of WordPiece tokenized category, attribute and product context. The outputs are token level attribute value spans in the product sources. We conduct inference on the cleaned Amazon product profiles using the five trained models. To ensure a high precision of the extractions, we further apply a simpler set of human extraction rules (created by human raters) to map the predicted spans to normalized attribute values, and remove the extractions that can not be mapped. The final extractions are aggregated from these five models into a positive set and a negative set as follows. For each example,
\begin{itemize}
    \item If the normalized attribute values from all five models are identical, the example is added to the positive set. The union of the extracted spans from all five models forms the span set of the attribute values.
    \item If no span is extracted from either model, and there is no extracted span from extraction rules, the example with empty value spans is put into the negative set, meaning no value for this attribute.
\end{itemize}
\begin{figure*}
\begin{center}
\includegraphics[width=0.8\linewidth]{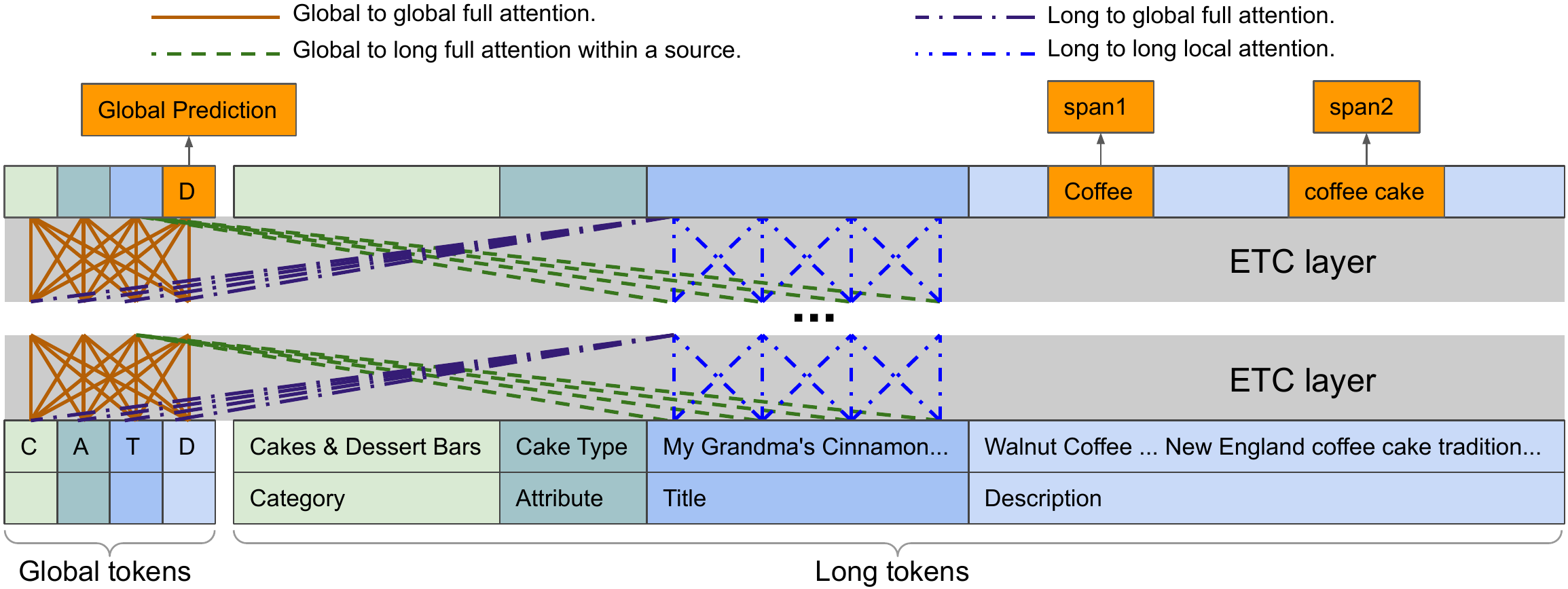}
\end{center}
\caption{Overall MAVEQA model architecture.} \label{fig:MAVEQA-model-architecture}
\end{figure*}
The aggregation ensures a low false positive rate in the positive set and a low false negative rate in the negative set. However, we still generated much more negative examples compared with positive examples. Therefore, we further down sample the negative examples by restricting each category-attribute pair having at most 5K negative examples.

The spans of the attribute values are very useful information in attribute value extraction research as well as questions answering systems. In this work, we map the token level attribute value spans to character level spans in the original sources following the method used in the BERT paper for the SQuAD task \cite{BERT}. If two spans overlap, we keep the one with the smallest begin index. However, in the MAVE dataset, we observe the spans are distributed sparsely with rare overlapping. The resulting (product, category, attribute, value spans) tuples are stored as an example in the MAVE dataset.


\subsection{Data Statistics}
The statistics of the MAVE dataset is reported in Table~\ref{table:general-statistics}. The original Amazon Reivew dataset contains 14.7M products. After the processing and filtering as described in Section~\ref{subsection:product-text-features} and \ref{subsection:generating-attribute-labels}, we have 2.2M products with 3M product-attribute annotations in the positive set, and 1.2M products with 1.8M product-attribute annotations in the negative set. In the positive set, the majority of products have one or two attributes, but there are still 100k+ products with more than three attributes.

\subsubsection{Multi-source and Sequence length statistics}
A product profile contains multiple sources, the distributions on number of sources and total number of words are shown in Figure~\ref{fig:sequence_length}, where a basic white space tokenizer is used to split the texts in all sources into words. It can be seen from the figures that the total number of words in a product is much longer than the typical number of words in a product title or description, which means the dataset contains much richer information from the multi-source product profiles. We observe that there is also a small but non-negligible portion of products (36k positives, 18k negatives) with sequence length longer than 512, which can serve as a data source for benchmarking models designed for long sequences.

\subsubsection{Attributes statistics}
The counts of unique categories, category-attribute pairs can be found in Table~\ref{table:general-statistics}. The average number of attributes per category is roughly 2, and the average number of categories an attribute shared across is 3. We visualize the number of positive example distributions on categories and attributes in Figure~\ref{fig:attribute-statistics}. It is clear that the head categories in general contain more attributes than average.

\section{Multi-source Attribute Value Extraction Model}\label{section:modeling}
The product profiles in MAVE dataset are composed of multiple text sources. Existing approaches might not work well on MAVE because: 1) They do not model the multi-source structure of the product, but simply treat product profile as one single text sequence by concatenating the texts from all sources. However, different source might carry different information, e.g., some attributes might appear more often in product title, while certain attributes are only mentioned in product specifications.
2) They are not able to deal with products with very long sequences. Long inputs are trimmed to fit into their models. In this work, to address these challenges, we propose a novel approach of Multi-source Attribute Value Extraction via Question Answering (MAVEQA), which effectively and efficiently models the products with structure and long profiles.

\subsection{Model Overview}
Following our previous work AVEQA \cite{WangYKSSSYE20}, we formulate the attribute value extraction task as a question answering problem. In particular, each attribute is treated as a question, and we seek the best answer span in the product context that corresponds to the value. 
The overall model architecture is shown in Figure \ref{fig:MAVEQA-model-architecture}. Essentially, our model consists of three main components, an input layer, a contextual encoder and an output layer. In the following sub-sections, we present each component
separately in detail.

\subsection{Input Layer}
Different from previous models ~\cite{XuWMJL19,WangYKSSSYE20,Amazon2}, our model treats each product source as an individual sequence, and assigns a global node/token to each of them. For example, in Figure \ref{fig:MAVEQA-model-architecture}, the product has two sources, i.e., one title and one description. By also regarding the attribute and category as two sources, our model assigns four global tokens to represent these four input sources respectively. The global token can be viewed as a summarization of its corresponding source. 
In the input layer, every word in each input source is converted into a $d$-dimensional embedding vector. This embedding is obtained by concatenating the word embedding and the source embedding. Each global token will also be initialized with a global token embedding.
Note that all the embeddings are trainable in our approach.

\subsection{Encoder}
In MAVEQA, we adopt the ETC encoder \cite{ETC} to generate the contextual embeddings for the global and long input tokens. ETC is an extension of the Transformer \cite{Attention} encoder to better capture the structure and long input, which is a natural fit to our task. The encoder in our model is a stack of identical ETC layers. There are four attention mechanisms in each ETC layer:
\begin{itemize}
    \item Global to global attention: each global token attends to all other global tokens, which allows information to flow among global tokens.
    \item Global to long attention: each global token attends to all long tokens in the corresponding source. Each global token is similar to a CLS token in the BERT model \cite{BERT}, which summarizes the corresponding source.
    \item Long to global attention: each long token in a source attends to all global tokens. This attention enables knowledge passing from other sources to the token in this source.
    \item Long to long attention: each long token will only attend to the long tokens within a local radius from the same source. This local attention pattern ensures the computational efficiency of the encoder.
\end{itemize}

\subsection{Output Layer}
In the output layer, a single dense layer with sigmoid activation is applied to the output long token embeddings, which computes a probability for every long token in all sources excluding category and attribute (as we are not seeking for values there) to predict whether the token is in an attribute value span. Similarly, the global token embeddings can be used to predict whether the corresponding source contains an attribute value or not. During training, we use sigmoid cross-entropy loss function on both long and global tokens.
\section{Experimental Results}
\begin{table}
\begin{adjustbox}{width=\columnwidth,center}
\begin{tabular}{|c|c|c|c|c|}
\hline
\multirow{2}{*}{\bf Attributes} &	\multicolumn{2}{c|}{\bf Train} & \multicolumn{2}{c|}{\bf Eval} \\
\cline{2-5}
&	\# positive & \# negative & \# positive & \# negative \\
\hline
\multicolumn{5}{|c|}{\bf Selected Head Attributes} \\
\hline
 Type & 805401 & 86661 & 100763 & 10738 \\
 Style & 138124 & 47899 & 17173 & 6151 \\
 Material & 93701 & 30204 & 11794 & 3762 \\
 Size & 21906 & 13323 & 2836 & 1690 \\
 Capacity & 15088 & 9149 & 1881 & 1161 \\
\hline
\multicolumn{5}{|c|}{\bf Selected Tail Attributes} \\
\hline
Black Tea Variety & 124 & 3434 & 15 & 439 \\
Staple Type & 135 & 249 & 22 & 33 \\
Web Pattern & 163 & 1110 & 19 & 144 \\
Cabinet Configuration & 158 & 112 & 24 & 15 \\
Power Consumption & 132 & 140 & 16 & 27 \\
\hline
\multicolumn{5}{|c|}{\bf All Attributes} \\
\hline
All Attributes & 2388842 & 1425936 & 298696 & 176978 \\
\hline
\multicolumn{5}{|c|}{\bf Zero-shot Attributes} \\
\hline
Special Occasion & 0 & 0 & 16142 & 105908 \\
Resolution & 0 & 0& 9509 & 5128 \\
Compatibility & 0 & 0& 6828 & 136 \\
Number of Sinks & 0 & 0& 349 & 446 \\
Food Processor Capacity & 0 & 0& 261 & 807 \\
\hline
\end{tabular}
\end{adjustbox}
\caption{Number of positive and negative examples in train and eval sets for the selected attributes, all attributes and zero-shot attributes benchmarks.}
\label{table:num-examples-split}
\end{table}

\subsection{Baseline models}\label{subsection:baseline-models}
We adopt the following models as baselines on the MAVE dataset.
\begin{itemize}
    \item \textbf{OpenTag} \cite{ZhengMD018} uses a BiLSTM-Attention-CRF architecture with sequence tagging strategies. OpenTag does not encode the attribute and thus builds one model per attribute. In order to scale up to an arbitrary number of attributes, similar to SUOpenTag model \cite{XuWMJL19}, we concatenate category, attribute, and source tokens in the input and make it attribute dependent. This attribute-dependent version of OpenTag is referred to {\bf ADOpenTag}.
    \item \textbf{AVEQA} \cite{WangYKSSSYE20} formulates the attribute value extraction as a question answering task. This model jointly encodes both attribute and product context with BERT encoder. It achieves the state-of-the-art results in the AE-110K dataset.
    \item \textbf{MAVEQA} is the new model proposed in this work, which adopts ETC \cite{ETC} encoder to encode structure and longer input sequences.
\end{itemize}
For all these baselines, we use the same output layer which is a single dense layer with sigmoid activation to predict token scores, and we use the same English uncased WordPiece vocabulary as in BERT \cite{BERT}. More implementation details are provided in the appendix.
\begin{table*}[t]
\small	
\begin{tabular}{|c|c|c|c|c|c|c|c|c|c|}
\hline
\multirow{2}{*}{\bf Attributes} & \multicolumn{3}{c|}{\bf (AD)OpenTag} & \multicolumn{3}{c|}{\bf AVEQA} & \multicolumn{3}{c|}{\bf MAVEQA} \\
\cline{2-10}
 & P (\%) & R (\%) & F (\%) & P (\%) & R (\%) & F (\%) & P (\%) & R (\%) & F (\%) \\
\hline
Type                 & 91.09 & 89.06 & 90.06 & 98.51  & 98.58  & {\bf 98.55} & 98.51 & 98.58 & {\bf 98.55} \\
Style                & 91.65 & 91.70 & 91.67 & 98.35  & 98.45  & 98.40 & 98.41 & 98.50 & {\bf 98.45} \\
Material             & 90.47 & 90.91 & 90.69 & 98.89  & 99.02  & {\bf 98.95} & 98.64 & 98.84 & 98.74 \\
Size                 & 80.32 & 79.44 & 79.88 & 96.95  & 97.60  & 97.28 & 97.32 & 97.50 & {\bf 97.41} \\
Capacity             & 86.94 & 86.39 & 86.67 & 98.09  & 98.35  & {\bf 98.22} & 98.19 & 98.19 & 98.19 \\
\hline
Black Tea Variety    & 100.00 & 80.00 & 88.89 & 100.00  & 100.00  & {\bf 100.00} & 100.00 & 100.00 & {\bf 100.00} \\
Staple Type          & 90.91 & 90.91 & 90.91 & 100.00 & 100.00 & {\bf 100.00} & 100.00 & 100.00 & {\bf 100.00} \\
Web Pattern          & 94.74 & 94.74 & 94.74 & 100.00  & 100.00  & {\bf 100.00} & 100.00 & 100.00 & {\bf 100.00} \\
Cabinet Configuration & 94.74 & 75.00 & 83.72 & 100.00  & 91.67  & {\bf 95.65} & 100.00 & 91.67 & {\bf 95.65} \\
Power Consumption    & 43.75 & 43.75 & 43.75 & 100.00  & 100.00  & {\bf 100.00} & 100.00 & 100.00 & {\bf 100.00} \\
\hline
All Attributes & 86.42 & 74.00 & 79.73 & 98.06 & 98.21 & 98.14 & 98.27 & 98.37 & {\bf 98.32} \\
\hline
\end{tabular}
\caption{Individual results on each selected attribute and average results on all attributes. {\bf OpenTag} is used for each selected attribute and {\bf ADOpenTag} is used for all attributes.} \label{table:random-split-and-selected}
\end{table*}

\subsection{Benchmarks}
We arrange the following setups for benchmark:
\begin{itemize}
    \item \textbf{Selected Attributes}
    To demonstrate the baseline model performance on individual attributes, we randomly select a set of attributes. In particular, we randomly select five attributes from the head ones that contain a large number of attribute-value annotations in the dataset. We also randomly select five tail attributes that have very few examples in the dataset. 
    \item \textbf{All Attributes}
    In order to evaluate the models' capability of scaling up to an arbitrary number of attributes, we also conduct experiments of all baselines on all attributes.
    \item \textbf{Zero-shot Attributes}
    To further examine the generalization ability of the models, we randomly select five attributes and holdout these attributes from training, i.e., none of these attributes are seen during training. We refer them to zero-shot attributes as they are used for evaluating zero-shot extraction.
    \item \textbf{Few-shot Learning}
    To study the impact of different number of training examples, we explore the model behavior with only a few training examples, namely few-shot learning.
\end{itemize}
We randomly split the dataset into train:eval:test = 8:1:1\footnote{We randomly sample up to 640k examples for training for faster convergence.}. Note that OpenTag trains a single model per attribute on the Selected Attributes. We use its extended version, {\bf ADOpenTag}, to train another model on all attributes.
The number of examples in each set is reported in Table~\ref{table:num-examples-split}.

\subsection{Metrics}
Following previous works ~\cite{XuWMJL19,WangYKSSSYE20,Amazon2}, we use Precision, Recall, and F1 score as evaluation metrics.
Specifically, for each attribute annotation, when the ground truth is No attribute value, the model can predict No value ({\bf NN}) or some incorrect Value ({\bf NV}); when the ground truth has attribute Values, the model can predict No value ({\bf VN}), Correct values ({\bf VC}) or Wrong values ({\bf VW}). We define {\bf NN}, {\bf NV}, {\bf VN}, {\bf VC}, and {\bf VW} as the number of examples for the above five cases. The precision and recall can be computed as:
\begin{equation*}
\begin{split}
{\bf P} = \frac{{\bf VC}}{{\bf NV} + {\bf VC} + {\bf VW}},\;\;\;
{\bf R} = \frac{{\bf VC}}{{\bf VN} + {\bf VC} + {\bf VW}}. \\
\end{split}
\end{equation*}
The {\bf F1} score is calculated as ${\bf 2PR}/({\bf P} + {\bf R})$.

\subsection{Results on Selected Attributes}
To study and compare the baseline model performances on individual attributes, we select five head attributes and five tail attributes. Note that an attribute can be shared across multiple categories. It can be seen from Figure~\ref{fig:attribute-statistics} that the number of categories containing the attribute and the number of the examples having the attribute are positively correlated. As aforementioned, we randomly select five head attributes, including {\it Type}, {\it Style}, {\it Material}, {\it Size}, and {\it Capacity}. It is clear that these attributes have more general meaning which are commonly shared across categories. The tail attributes have a small number of examples and often only appear in 1-2 categories. In order to have a meaningful evaluation on tail attributes, we also constrain the number of evaluation examples to be larger than 15. The tail attributes include {\it Black Tea Variety}, {\it Staple Type}, {\it Web Pattern}, {\it Cabinet Configuration} and {\it Power Consumption}. 
\begin{figure}
\centering
\includegraphics[width=0.82\linewidth]{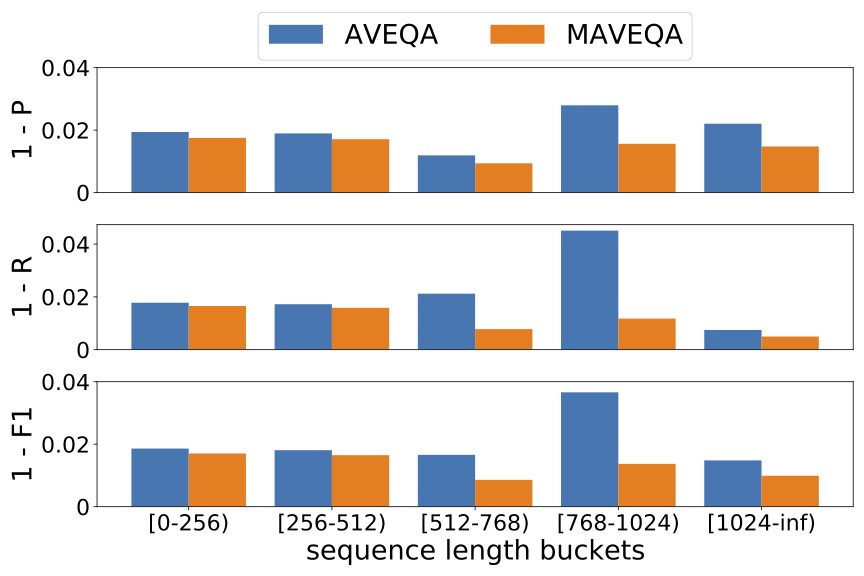}
\caption{Precision, Recall, F1 errors within each bucket of example sequence length for all attributes.} \label{fig:sequence-length-bucket}
\end{figure}

The results of the OpenTag, AVEQA, and MAVEQA models on these selected attributes are reported in Table~\ref{table:random-split-and-selected}. From these comparison results, we can see that the AVEQA and MAVEQA models perform consistently better than OpenTag on all head and tail attributes. The reason is that both AVEQA and MAVEQA use the advanced Transformer architecture which have been shown to outperform BiLSTM-Attention-CRF architecture. Moreover, the knowledge can be transferred among different attributes in AVEQA and MAVEQA, since they jointly encode the attribute and the product across all attributes. 
It also can be seen that AVEQA and MAVEQA achieve similar results on these selected attributes. We found out, by examining the evaluation examples, that the product sequences of these selected attributes are not long, and thus AVEQA and MAVEQA are equally effective when modeling short sequences. We will discuss more on the effect of product sequence length in the next section. Another interesting observation is that the results of MAVEQA on the tail attributes are even better (F1 scores are almost 100\%) than those on the head attributes. This is because although tail attributes have relatively small training examples, their attribute-value patterns are much simpler than those in the head attribute. For example, {\it Black Tea Variety} is a specific attribute in category {\it Tea $\&$ Infusions}, which only has three possible values, i.e. {\it Ceylon}, {\it Darjeeling} and {\it Earl Grey}. On the other hand, the attribute {\it Type} appears in more than 40 categories with more than 500 different values. However, MAVEQA is still able to achieve very high performance on the head attributes.

\subsection{Results on All Attributes}
The average results over all attributes are reported in Table~\ref{table:random-split-and-selected}. We observe similar result patterns to those in the selected attributes. It can be seen that MAVEQA performs slightly better than AVEQA. Our hypothesis is that for those examples with multiple product sources and long sequences, MAVEQA is more effective in capturing the attribute-value relations via modeling the complete structure and long sequence. In contrast, AVEQA simply concatenates and trims the text from different sources and treats it as a single sequence. 
To further investigate along this direction, we divide the evaluation examples into different buckets w.r.t. the sequence length of the example, and compute the metrics in each bucket for AVEQA and MAVEQA. The precision, recall and F1 gap/error results are shown in Figure~\ref{fig:sequence-length-bucket}. It can be seen that with example sequence length exceeding max sequence length of the models, both precision and recall errors for AVEQA go up, while the errors remain the same for MAVEQA. This observation validates that MAVEQA is indeed more effective on examples with longer sequences.
\begin{table}
\begin{adjustbox}{width=\columnwidth,center}
\begin{tabular}{|c|c|c|c|c|c|c|}
\hline
\multirow{2}{*}{\bf Zero-shot Attributes} &\multicolumn{3}{c|}{\bf AVEQA} & \multicolumn{3}{c|}{\bf MAVEQA} \\
\cline{2-7}
& P (\%) & R (\%) & F (\%) & P (\%) & R (\%) & F (\%)\\
\hline
Special Occasion     & 62.13 & 50.16 & {\bf 55.51} & 10.06 & 0.80 & 1.48 \\
Resolution           & 80.95 & 38.55 & 52.23 & 85.56 & 44.94 & {\bf 58.93} \\
Compatibility        & 0.00 & 0.00 & 0.00 & 0.00 & 0.00 & 0.00 \\
Number of Sinks     & 39.10 & 29.80 & 33.82 & 48.57 & 38.97 & {\bf 43.24} \\
Food Processor Capacity& 81.97 & 92.34 & 86.85 & 83.50 & 98.85 & {\bf 90.53} \\
\hline
\end{tabular}
\end{adjustbox}
\caption{Results on zero-shot attributes.} \label{table:holdout}
\end{table}

\subsection{Results on Zero-shot Attributes}\label{subsection:zero-shot-results}
We conduct zero-shot extraction experiments to evaluate the generalization ability of AVEQA and MAVEQA on unseen attributes. The zero-shot extraction results are reported in Table~\ref{table:holdout}. There are several interesting observations from this table. First, it is clear that the overall performance of both methods on these zero-shot attributes are much worse compared with those results in Table~\ref{table:random-split-and-selected}, which is consistent with our expectation as there are no training examples for the zero-shot attributes. Second, AVEQA achieves better results than MAVEQA on {\it Special Occasion}. Our hypothesis is that AVEQA and MAVEQA are trained with different pre-trained models, i.e., AVEQA is initialized with the pre-trained BERT \cite{BERT} while MAVEQA is initialized from ETC \cite{ETC}. Previous works ~\cite{KDL,LWF,WangYKSSSYE20} have shown that the pre-trained model is crucial for the fine-tuned model to perform well on zero-shot learning. However, the pre-trained ETC model removes all examples with less than 10 sentences since it focuses on long sequences, which might have eliminated a certain amount of knowledge on attribute {\it Special Occasion}, and thus does not perform well. Finally, AVEQA and MAVEQA achieve 0 scores in terms of all metrics on {\it Compatibility}, which indicates that it is a truly difficult zero-shot attribute. Both pre-trained models and the training examples of other attributes are lacking information about {\it Compatibility}, resulting in 0 extractions. We believe there is deep knowledge and information that are uncovered in zero-shot extraction. The MAVE dataset can serve as a rich resource to drive research in this direction.


\subsection{Results on Few-shot Learning}
\begin{figure}
\centering
\includegraphics[width=0.83\linewidth]{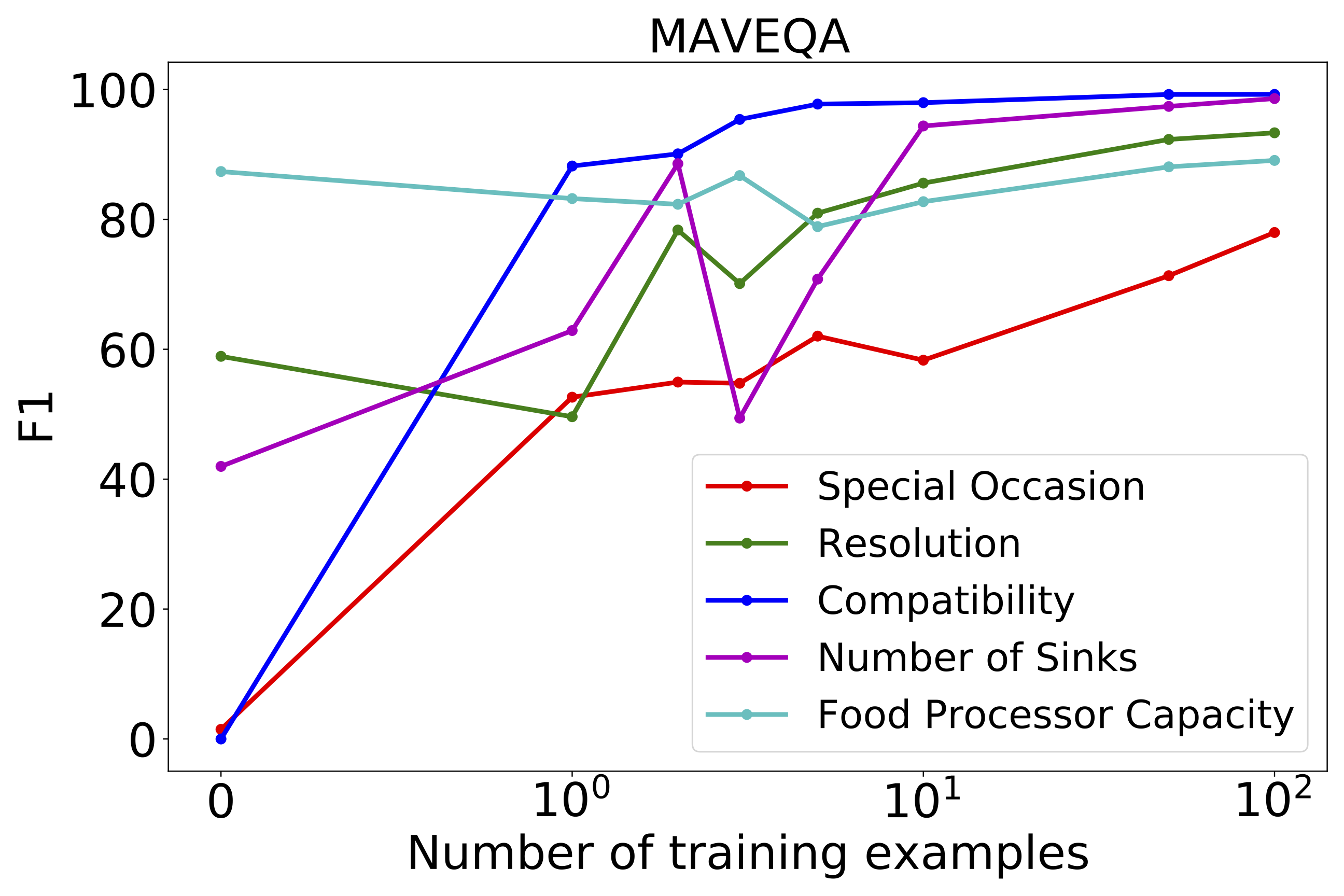}
\caption{Results on Few-shot learning.} \label{fig:few-shot-maveqa}
\end{figure}

To study the effectiveness of MAVEQA with few training examples, we conduct few-shot experiments by varying the number of training examples, $k$, from \{1, 2, 3, 5, 10, 50, 100\} on all zero-shot attributes. For each run, we first randomly separate out 100 examples per attribute, then we randomly select $k$ examples from the 100 examples. We perform fine-tuning for 20K steps over these selected examples initializing from the model trained on all attributes (exclude zero-shot attributes). The F1 scores of MAVEQA on few-shot experiments are shown in Figure~\ref{fig:few-shot-maveqa}. It is not surprising to see that the F1 scores increase with more training examples. We can also observe that the F1 scores on most attributes are able to reach a reasonably high value, above 90\%, when the number of training examples increases to 100. We notice that the F1 score remains flattened for attribute {\it Food Processor Capacity}. By analyzing the errors, we found that there are a certain amount of false negatives in the evaluation set, where the extractions are actually correct but the attribute value annotations are missing. Improving the attribute coverage on MAVE is definitely one of the future directions.

\section{Conclusion}
In this paper, we introduce the Multi-source Attribute Value Extraction (MAVE) dataset – the largest, multi-source, diverse, context-rich, clean dataset. 
By extracting and verifying attribute values from over 2.2 million multi-source product profiles, MAVE provides a large set of 3 million attribute-value annotations. 
We establish benchmarks on MAVE with several state-of-the-art methods, including a novel approach proposed in this work that models the multi-source and long product profiles.
We also empirically demonstrated the use of this dataset on a very challenging task - zero-shot attribute value extraction, which is not fully exploited by existing datasets.
We believe this can serve as a rich resource to drive research in the attribute value extraction domain for years to come and enable the community to build better and more generalized models.
MAVE can potentially be used as a pretraining dataset in various downstream tasks, including product embedding learning, attribute grounded questions answering, product search and recommendation, etc. 

{\bibliographystyle{abbrv}
\footnotesize{\bibliography{references}}}

\appendix

\section{Appendix}
\subsection{Implementation Details}
For all these baselines, we use the same output layer which is a single dense layer with sigmoid activation to predict token scores, and we use the same English uncased WordPiece vocabulary as in BERT. For {\bf ADOpenTag} and {\bf AVEQA}, we use the input format of "[CLS] category tokens [SEP] attribute tokens [SEP] all source tokens [SEP]". For {\bf OpenTag} and {\bf MAVEQA}, we do not use any special tokens such as [CLS], [SEP]. This is to balance between minimizing the difference of the input formats and better utilizing pretrained models.
Table~\ref{table:hyper-parameters} contains the hyper-parameters details for training the {\bf (AD)OpenTag}, {\bf AVEQA}, and {\bf MAVEQA} models.

\begin{table}[h!]
\small
\begin{tabular}{l|c}
\hline
{\bf Hyper-parameters} & {\bf Value} \\
\hline
{\bf Common} & \\
batch size & 128 \\
iterations & 200,000 \\
optimizer & Adam \\
Adam $\beta_1$, $\beta_1$, $\epsilon$ & 0.9, 0.99, $1e^{-6}$ \\ 
learning rate schedule & linear decay \\
initial learning rate & $3e^{-5}$ \\
end learning rate & 0.0 \\
learning rate warmup steps & 10,000 \\
vocab size & 30522 \\
\hline
{\bf (AD)OpenTag} & \\
max sequence length & 512 \\
word embedding size & 768 \\
unidirectional LSTM units & 512 \\
LSTM inputs dropout & 0.0 \\
LSTM recurrent dropout & 0.4 \\
attention dropout & 0.0 \\
word embedding layer init & \href{https://tfhub.dev/tensorflow/bert_en_uncased_L-12_H-768_A-12/4}{TFHub BERT-base} \\
\hline
{\bf AVEQA} & \\
max sequence length & 512 \\
transformer layers & 12 \\
transformer attention heads & 12 \\
transformer hidden dimension & 768 \\
pretrained checkpoint & \href{https://tfhub.dev/tensorflow/bert_en_uncased_L-12_H-768_A-12/4}{TFHub BERT-base} \\
\hline
{\bf MAVEQA} & \\
max long sequence length & 1024 \\
max global sequence length & 64 \\
transformer layers & 12 \\
transformer attention heads & 12 \\
transformer hidden dimension & 768 \\
local attention radius & 84 \\
pretrained checkpoint & \href{https://github.com/google-research/google-research/tree/master/etcmodel#pre-trained-models}{etc\_base\_2x\_pretrain} \\
\hline
\end{tabular}
\caption{Model Hyper-parameters details.}
\label{table:hyper-parameters}
\end{table}

\begin{table}[H]
\small
\begin{tabular}{c|c|c|c}
\hline
{\bf Models} & {\bf \# params} & {\bf Hardware} & {\bf Training time} \\
\hline
{\bf OpenTag} & 28.7M & 32 core TPU v3 & 7.1h  \\
{\bf AVEQA} & 109M & 32 core TPU v3 & 2.7h  \\
{\bf MAVEQA} & 166M & 64 core TPU v3 & 7.2h  \\
\hline
\end{tabular}
\caption{Model parameters and training hardware.}
\label{table:parameters-hardware-time}
\end{table}

\subsection{Number of Model Parameters, Hardware and Training Cost}
The number of model parameters, hardwares to train the model and training time are shown in Table~\ref{table:parameters-hardware-time}. The larger number of parameters for {\bf MAVEQA} is due to the parameters for global nodes in the ETC model. Since {\bf MAVEQA} uses a longer sequence length and has more parameters, we use more TPU cores to train. We use Tensorflow distribute strategy with SUM loss reduction. We note that this is not an issue since we are using Adam optimizer which normalizes the gradients by the second raw momentum estimate.

Note that even {\bf (AD)OpenTag} has much smaller number of parameters than {\bf AVEQA}, it takes much longer time to train on TPUs, which may due to the recurrent nature of its LSTM component. {\bf MAVEQA} takes much longer than {\bf AVEQA}, which may due to a longer sequence length and the implementation of ETC's local attention on TPU being not as efficient as full attention.

\subsection{Additional Results on Few-Shot Learning}
\begin{figure}[h]
\centering
\includegraphics[width=0.82\linewidth]{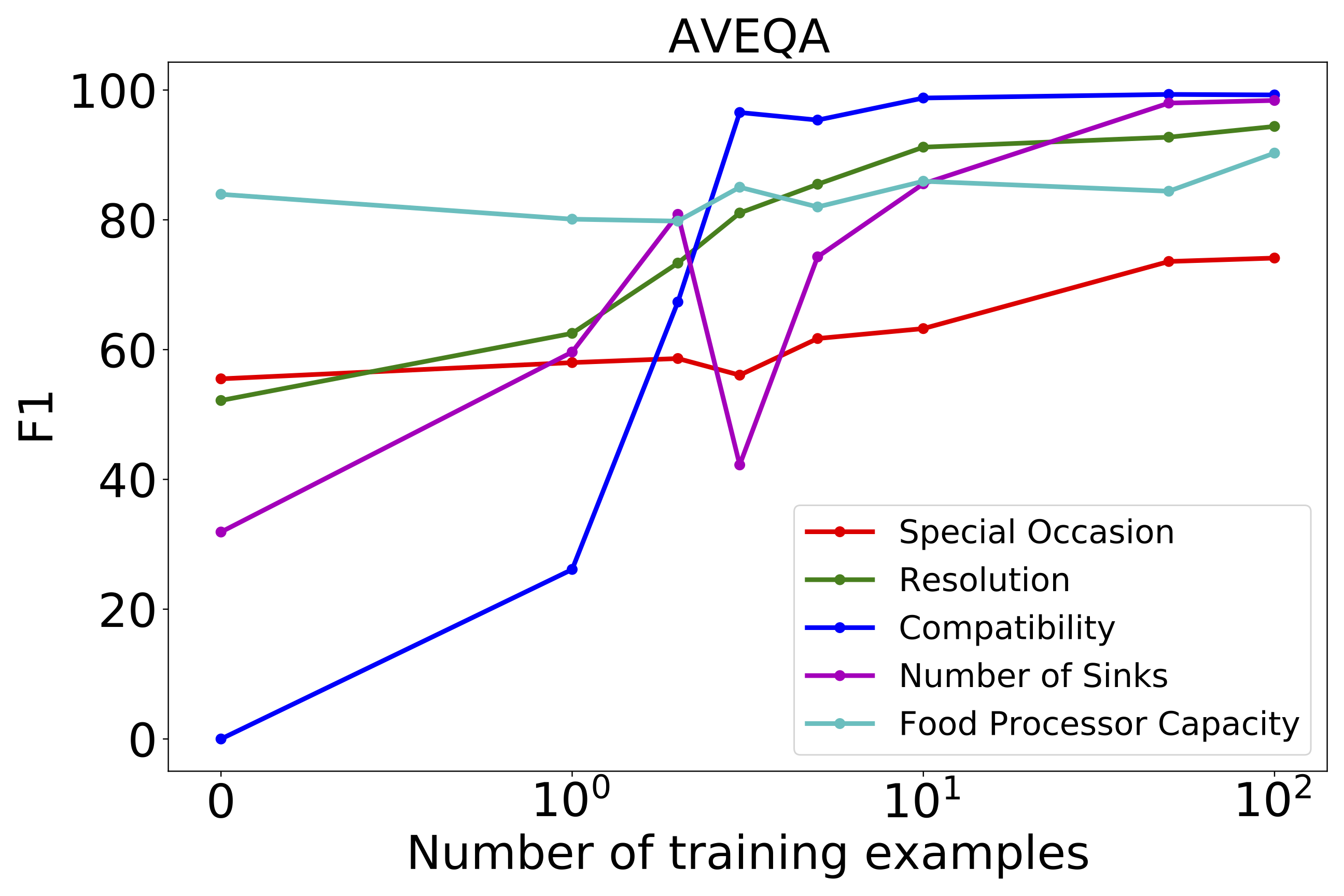}
\caption{Impact of different numbers of training examples for the {\bf AVEQA} model.} \label{fig:few-shot-aveqa}
\end{figure}

We provide more results on the few-shoot learning using {\bf AVEQA} in Figure~\ref{fig:few-shot-aveqa}, which has a similar trend as {\bf MAVEQA}.

\subsection{Ablation Study}
We also used different sizes of BERT backbones for {\bf AVEQA}. The BERT parameters are initialized from TFHub pretrained checkpoints. The results on random split of all attributes are shown in Table~\ref{table:bert-ablation}. We can see that as model size increases, the results constantly become better. However, the BERT-large results are just close to the {\bf MAVEQA} results in Table~5 in the main text, which suggests the advantage of {\bf MAVEQA} over {\bf AVEQA}.

\begin{table}[H]
\small
\begin{tabular}{|l|r|r|r|}
\hline
BERT sizes & P (\%) & R (\%) & F (\%) \\
\hline
\href{https://tfhub.dev/tensorflow/small_bert/bert_en_uncased_L-2_H-128_A-2/2}{BERT-2L} & 91.60 & 90.14 & 90.87 \\
\href{https://tfhub.dev/tensorflow/small_bert/bert_en_uncased_L-4_H-256_A-4/2}{BERT-4L} & 96.78 & 96.92 & 96.85 \\
\href{https://tfhub.dev/tensorflow/small_bert/bert_en_uncased_L-6_H-512_A-8/2}{BERT-6L} & 97.98 & 98.17 & 98.07 \\
\href{https://tfhub.dev/tensorflow/small_bert/bert_en_uncased_L-8_H-768_A-12/2}{BERT-8L} & 98.10 & 98.27 & 98.18 \\
\href{https://tfhub.dev/tensorflow/bert_en_uncased_L-12_H-768_A-12/4}{BERT-base} & 98.06 & 98.21 & 98.14 \\
\href{https://tfhub.dev/tensorflow/bert_en_uncased_L-24_H-1024_A-16/4}{BERT-large} & {\bf 98.29} & {\bf 98.33} & {\bf 98.31} \\
\hline
\end{tabular}
\caption{Ablation study of different sizes of BERT backbone in {\bf AVEQA}. The metrics are for random split on all attributes.} 
\label{table:bert-ablation}
\end{table}

\end{document}